\documentclass[runningheads]{llncs}
\usepackage{orcidlink}

\usepackage{graphicx}
\usepackage{multirow}
\usepackage{booktabs}
\usepackage{tikz}
\usepackage{comment}
\usepackage{amsmath,amssymb} 
\usepackage{color}
\usepackage{hyperref} 
\hypersetup{
    colorlinks=true,
    linkcolor=red,
    urlcolor=magenta,
    linktoc=all,
    citecolor=green
           }
\usepackage[accsupp]{axessibility}  

\newcommand{\Figref}[1]{Fig.\ref{#1}}

\newcommand{\tabref}[1]{Tab.\ref{#1}}
\newcommand{\eqnref}[1]{Eq.(\ref{#1})}
\newcommand{\secref}[1]{Sec.\ref{#1}}

\newcommand{\etal}{\textit{et al.}}

\begin{document}

\pagestyle{headings}
\mainmatter
\def\ECCVSubNumber{19}  

\title{Learning Depth from Focus in the Wild} 

%
\author{Changyeon Won\orcidlink{0000-0001-5335-2606}\and
Hae-Gon Jeon\orcidlink{0000-0003-1105-1666}\thanks{Corresponding author}}
\authorrunning{Won and Jeon}
%
\institute{Gwangju Institute of Science and Technology \\\email{cywon1997@gm.gist.ac.kr} and \email{haegonj@gist.ac.kr}}

\maketitle

\begin{abstract}
   For better photography, most recent commercial cameras including smartphones have either adopted large-aperture lens to collect more light or used a burst mode to take multiple images within short times. These interesting features lead us to examine depth from focus/defocus. 
   In this work, we present a convolutional neural network-based depth estimation from single focal stacks. Our method differs from relevant state-of-the-art works with three unique features. First, our method allows depth maps to be inferred in an end-to-end manner even with image alignment. Second, we propose a sharp region detection module to reduce blur ambiguities in subtle focus changes and weakly texture-less regions. Third, we design an effective downsampling module to ease flows of focal information in feature extractions. In addition, for the generalization of the proposed network, we develop a simulator to realistically reproduce the features of commercial cameras, such as changes in field of view, focal length and principal points.
  By effectively incorporating these three unique features, our network achieves the top rank in the DDFF 12-Scene benchmark on most metrics. We also demonstrate the effectiveness of the proposed method on various quantitative evaluations and real-world images taken from various off-the-shelf cameras compared with state-of-the-art methods. Our source code is publicly available at \url{https://github.com/wcy199705/DfFintheWild}.

\end{abstract}

\section{Introduction}

  \begin{figure}[t]
    \includegraphics[width=1\linewidth]{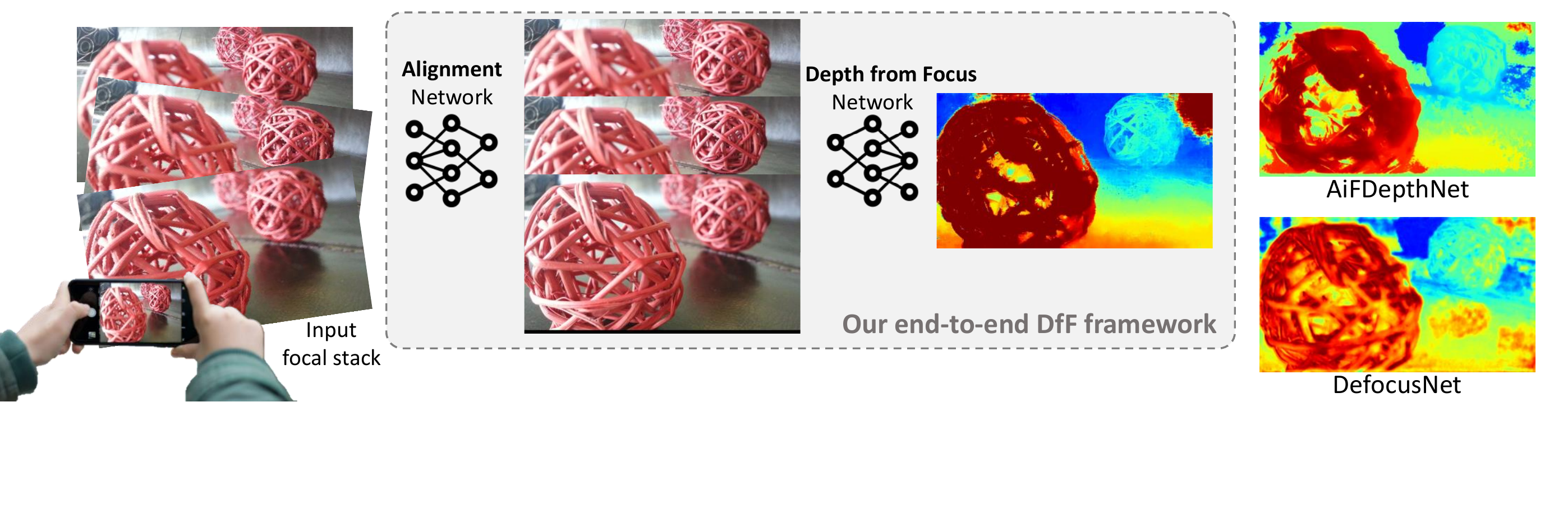}
    \vspace{-1.5cm}
    \caption{Results of our true end-to-end DfF framework with comparisons to state-of-the-art methods.} 
    \label{fig:teaser}
    \vspace{-0.3cm}
\end{figure}

 As commercial demand for high-quality photographic applications increases, images have been increasingly utilized in scene depth computation.
 Most commercial cameras, including smartphone and DSLR cameras have two interesting configurations: large-aperture lens and a dual-pixel (DP) sensor. 
 Both are reasonable choices to collect more light and to quickly sweep the focus through multiple depths. Because of this, images appear to have a shallow depth of field (DoF) and are formed as focal stacks with corresponding meta-data such as focal length and principal points. 
  One method to accomplish this is to use single dual-pixel (DP) images which have left and right sub-images with narrow baselines and limited DoFs. A straightforward way is to find correspondences between the left and right sub-images~\cite{wadhwa2018synthetic,garg2019learning,zhang20202}. Despite an abundance of research, such methods are heavily dependent on the accurate retrieval of correspondences due to the inherent characteristics of DP images. Pixel disparities between the two sub-images result in blurred regions, and the amount of spatial shifts is proportional to the degree of blurrings.
 Another group of approaches solves this problem using different angles. The out-of-focus regions make it possible to use depth-from-defocus (DfD) techniques to estimate scene depths~\cite{anwar2017depth,pan2021dual,zhang2021joint}. 
 Since there is a strong physical relationship between scene depths and the amount of defocus blurs, the DfD methods account for it in data-driven manners by learning to directly regress depth values.
 However, there is a potential limitation to these works~\cite{anwar2017depth,pan2021dual,zhang2021joint}. A classic issue, an  
 aperture effect, makes an analysis of defocus blur in a local window difficult. In addition, some of them recover deblurred images from input, but image deblurring also belongs to a class of ill-posed inverse problems for which the uniqueness of the solution cannot be established~\cite{levin2007image}.

These shortcomings motivate us to examine depth from focus (DfF) as an alternative. DfF takes in a focal stack to a depth map during focus sweeping, which is available in most off-the-shelf cameras, and determines the focus in the input focal stack. In particular, the inherent operations of convolutional neural networks (CNNs), convolution and maxpooling, are suitable for measuring the values obtained from derivatives of the image/feature map based on the assumption that focused images contain sharper edges~\cite{hazirbas2018deep,maximov2020focus,Wang-ICCV-2021}. Nevertheless, there is still room for improvements with respect to model generalization, due to the domain gap between public datasets and real-world focal stack images, and an alignment issue that we will discuss. 
 
 In this work, we achieve a high-quality and well-generalized depth prediction from single focal stacks. Our contributions are threefold (see \Figref{fig:teaser}): First, we compensate the change in image appearance due to magnification during the focus change, and the slight translations from principal point changes. Compared to most CNN-based DfD/DfF works~\cite{hazirbas2018deep,maximov2020focus,Wang-ICCV-2021} which either assume that input sequential images are perfectly aligned or use hand-crafted feature-based alignment techniques, we design a learnable context-based image alignment, which works well in defocusing blurred images. Second, the proposed sharp region detection (SRD) module addresses blur ambiguities resulting from subtle defocus changes in weakly-textured regions. SRD consists of convolution layers and a residual block, and allows the extraction of more powerful feature representations for image sharpness. Third, we also propose an efficient downsampling (EFD) module for the DfF framework. The proposed EFD combines output feature maps from upper scales using a stride convolution and a 3D convolution with maxpooling and incorporates them to both keep the feature representation of the original input and to ease the flow of informative features for focused regions. To optimize and generalize our network, we develop a high performance simulator to produce photo-realistic focal stack images with corresponding meta-data such as camera intrinsic parameters. 
 
 With this depth from focus network, we achieve state-of-the-art results over various public datasets as well as the top rank in the DDFF benchmark~\cite{hazirbas2018deep}. Ablation studies indicate that each of these technical contributions appreciably improves depth prediction accuracy.

\section{Related Work}
The mainstream approaches for depth prediction such as monocular depth estimation~\cite{fu2018deep,godard2017unsupervised,godard2019digging}, stereo matching~\cite{chang2018pyramid,shen2021cfnet} and multiview stereo~\cite{gu2020cascade,im2019dpsnet} use all-in-focus images. As mentioned above, they overlook the functional properties of off-the-shelf cameras and are out-of-scope for this work. In this section, we review depth from defocus blur images, which are closely related to our work.

\noindent\textbf{Depth from Defocus.}\quad
 Some unsupervised monocular depth estimation~\cite{srinivasan2018aperture,gur2019single} approaches utilize a defocus blur cue as a supervisory signal. A work in~\cite{srinivasan2018aperture} proposes differentiable aperture rendering functions to train a depth prediction network which generates defocused images from input all-in-focus images. The network is trained by minimizing distances between ground truth defocused images and output defocused images based on an estimated depth map. Inspired by~\cite{srinivasan2018aperture}, a work in~\cite{gur2019single} introduces a fast differentiable aperture rendering layer from hypothesis of defocus blur. In spite of depth-guided defocus blur, both these works need all-in-focus images as input during an inference time. Anwar~\etal~\cite{anwar2017depth} formulate a reblur loss based on circular blur kernels to regularize depth estimation, and design a CNN architecture to minimize input blurry images and reblurred images from output deblurring images as well.
 Zhang and Sun~\cite{zhang2021joint} propose a regularization term to impose a consistency between depth and defocus maps from single out-of-focus images.

 \noindent\textbf{Depth from DP images.}\quad

Starting with the use of traditional stereo matching, CNN-based approaches have been adopted for depth from DP images~\cite{wadhwa2018synthetic}. A work in~\cite{garg2019learning} introduces that an affine ambiguity exists between a scene depth and its disparity from DP data, and then alleviates it using both novel 3D assisted loss and folded loss. In~\cite{zhang20202}, a dual-camera with DP sensors is proposed to take advantage of both stereo matching and depth from DP images. In~\cite{punnappurath2020modeling},  unsupervised depth estimation by modeling a point spread function of DP cameras. The work in~\cite{pan2021dual} proposes an end-to-end CNN for depth from single DP images using both defocus blur and correspondence cues. In addition, they provide a simulator that makes a synthetic DP dataset from all-in-focus images and the corresponding depth map. In \cite{xin2021defocus}, single DP images are represented via multi-plane images \cite{tucker2020single} with a calibrated point spread function for a certain DP camera model. The representation is used for both unsupervised defocus map and all-in-focus image generation.

\noindent\textbf{Depth from Focus.}\quad
 DfF accounts for changes in blur sizes in the focal stack and determines scene depths according to regions adjacent to the focus~\cite{pertuz2013analysis,levin2007image,maximov2020focus}. In particular, conventional DfF methods infer depth values from a focal stack by comparing the sharpness of a local window at each pixel~\cite{jeon2019ring,sakurikar2017composite,suwajanakorn2015depth}. 
The research in~\cite{hazirbas2018deep} introduces a CNN-based DfF by leveraging focal stack datasets made with light field and RGB-D cameras. In~\cite{maximov2020focus}, domain invariant defocus blur is used for dealing with the domain gap. The defocus blur is supervised to train data-driven models for DfF as an intermediate step, and is then utilized for permutation-invariant networks to achieve a better generalization from synthetic datasets to real photos. In addition, the work uses a recurrent auto-encoder to handle scene movements which occur during focal sweeps\footnote{Unfortunately, both the source codes for training/test and its pre-trained weight are not available in public.}. In~\cite{Wang-ICCV-2021}, a CNN learns to make an intermediate attention map which is shared to predict scene depth prediction and all-in-focus images reconstruction from focal stack images.

  \begin{figure}[t]
    \includegraphics[width=1\linewidth]{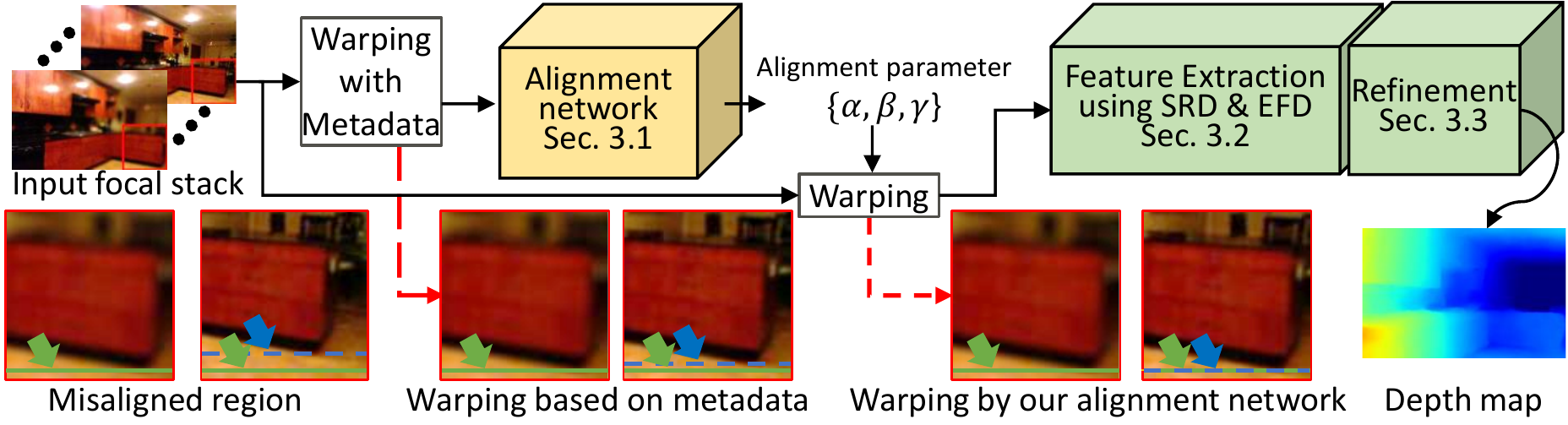}
    \vspace{-0.7cm}
    \caption{An overview of the proposed network.} 
    \label{fig:overview}
    \vspace{-0.2cm}
\end{figure}

\section{Methodology}
Our network is composed of two major components: One is an image alignment model for sequential defocused images. It is a prerequisite that we should first address the non-alignment issue on images captured with smartphones whose focus is relayed to focus motors adjusting locations of camera lenses.
Another component is a focused feature representation, which encodes the depth information of scenes. 
To be sensitive to subtle focus changes, it requires two consecutive feature maps of the corresponding modules from our sharp region detector (SRD) and an effective downsampling module for defocused images (EFD). The overall procedure is depicted in~\Figref{fig:overview}.

\subsection{A Network for Defocus Image Alignment}
\label{sec:alignment}
 Since camera field of views (FoVs) vary according to the focus distance, a zoom-like effect is induced during a focal sweep~\cite{herrmann2020learning}, called focal breathing. Because of the focal breathing, an image sharpness cannot be accurately measured on the same pixel coordinates across focal slices. As a result, traditional DfF methods perform  feature-based defocus image alignment to compensate this, prior to depth computations. However, recent CNN-based approaches disregard the focal breathing because either all public synthetic datasets for DfF/DfD, whose scale is enough to generalize CNNs well, provide well-aligned focal stacks, or are generated by single RGB-D images. 
 Because of this gap between real-world imagery and easy to use datasets, their generality is limited. Therefore, as a first step to implementing a comprehensive, all-in-one solution to DfF, we introduce a defocus image alignment network.

 \noindent\textbf{Field of view.}\quad
 Scene FoVs are calculated by work distances, focus distances, and the focal length of cameras in \eqnref{eq:Fov}. Since the work distances are fixed during a focal sweep, relative values of FoVs (Relative FoVs) are the same as the inverse distance between sensor and lens. We thus perform an initial alignment of a focal stack using these relative FoVs as follows:
 \begin{gather}
 \label{eq:Fov}
     FoV_{n} = W \times \frac{A}{s_{n}}\\
     Relative \ FoV_{n} = \frac{FoV_{n}}{FoV_{min}} =  \frac{s_{min}}{s_{n}} \notag
    \ \ (s_{n} =   \frac{F_{n}\times f}{F_{n}-f})\notag,\\
 \notag
 \end{gather} 
 where $s_n$ is the distance between the lens and the sensor in an $n$-$th$ focal slice. $A$ is the sensor size, and $W$ is the working distance. $f$ and $F_{n}$ are the focal length of the lens and a focus distance, respectively.
 $min$ denotes an index of a focal slice whose FoV is the smallest among focal slices. In this paper, we call this focal slice with the $min$ index as the target focal slice. We note that the values are available by accessing the metadata information in cameras without any user calibration.

 Nevertheless, the alignment step is not perfectly appropriate for focal stack images due to hardware limitations, as described in~\cite{herrmann2020learning}. Most smartphone cameras control their focus distances by spring-installed voice coil motors (VCMs). The VCMs adjust the positions of the camera lens by applying voltages to a nearby electromagnet which induces spring movements. Since the elasticity of the spring can be changed by temperature and usage, there will be an error between real focus distances and values in the metadata. In addition, the principal point of cameras also changes during a focal sweep because the camera lens is not perfectly parallel to the image sensor, due to some manufacturing imperfections. Therefore, we propose an alignment network to adjust this mis-alignment and a useful simulator to ensure realistic focal stack acquisition.
  \begin{figure}[t]
  \centering
    \includegraphics[width=1\linewidth]{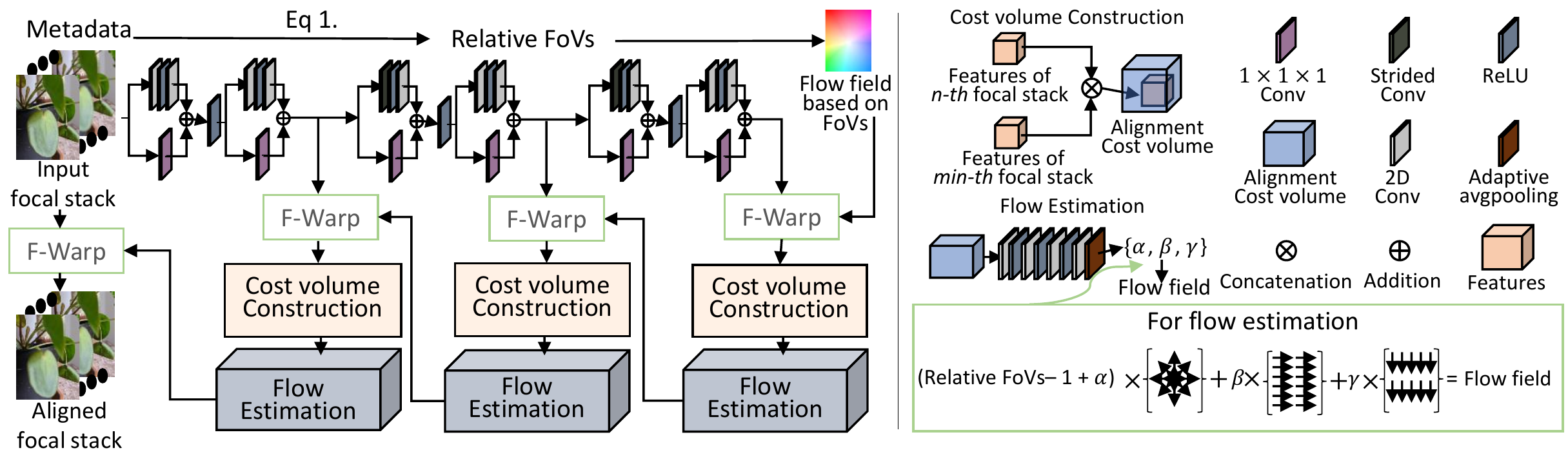}
    \vspace{-0.7cm}
    \caption{An illustration of our alignment network. Given initially-aligned images with camera metadata, this network produces an aligned focal stack. In the flow estimation, we use three basis functions to model radial, horizontal and vertical motions of VCMs.}
 \label{fig:123}
 \vspace{-0.2cm}
\end{figure}

 \noindent\textbf{Alignment network.}\quad
 As shown in~\Figref{fig:123}, our alignment network has 3-level encoder-decoder structures, similar to the previous optical flow network~\cite{hui2018liteflownet}. The encoder extracts multi-scale features, and multi-scale optical flow volumes are constructed by concatenating the features of a reference and a target focal slice. The decoder refines the multi-scale optical flow volumes in a coarse-to-fine manner using feature warping (F-warp).
  However, we cannot directly use the existing optical flow framework for alignment because defocus blur breaks the brightness constancy assumption~\cite{suwajanakorn2015depth}. 
  
  To address this issue, we constrain the flow using three basis vectors with corresponding coefficients ($\alpha,~\beta,~\gamma$) for each scene motion. To compute the coefficients instead of the direct estimation of the flow field, we add an adaptive average pooling layer to each layer of the decoder. The first basis vector accounts for an image crop which reduces errors in the FoVs. We elaborate the image crop as a flow that spreads out from the center.
  The remaining two vectors represent $x-$ and $y-$axis translations, which compensate for errors in the principal point of the cameras. These parametric constraints of flow induce the network to train geometric features which are not damaged by defocus blur. We optimize this alignment network using a robust loss function $L_{align}$, proposed in~\cite{liu2019ddflow}, as follows:

  \begin{equation}
\label{eq:alignment loss}
  L_{align} = \sum_{n=0}^N \rho ( I_{n}( \Gamma + D(\Gamma) ) - I_{min} (\Gamma)),
  \end{equation}
  where $ \rho(\cdot) = ( |\cdot| + \varepsilon )^q$. $q$ and $\varepsilon$ are set to 0.4 and 0.01, respectively. $I_{n}$ is a focal slice of a reference image, and $I_{min}$ is the target focal slice. $D(\Gamma)$ is an output flow of the alignment network at a pixel position, $\Gamma$.
  
  We note that the first basis might be insufficient to describe the zooming effects with spatially-varying motions. However, our design for the image crop shows consistently promising results for the alignment, thanks to the combination of the three basis functions that compensate for a variety of motions in real-world.
  \begin{figure}[ht]
    \includegraphics[width=1\linewidth]{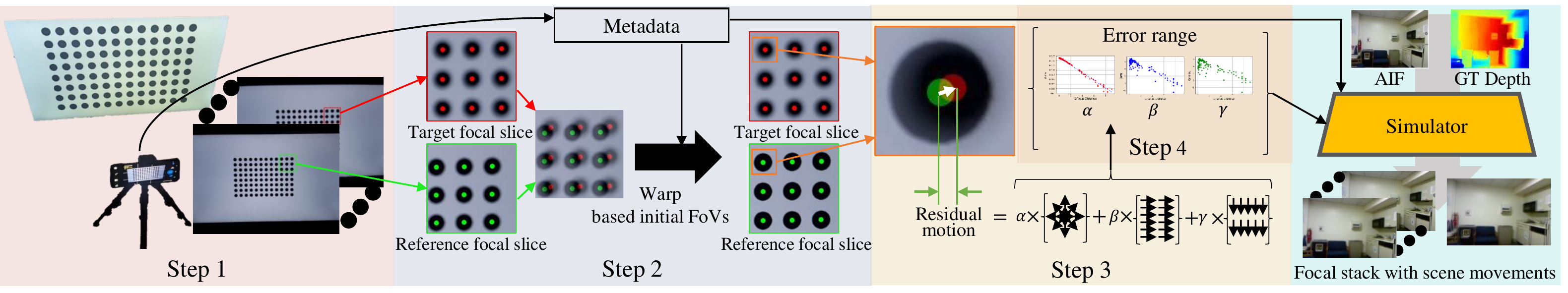}
    \vspace{-0.9cm}
    \caption{A pipeline of our simulator. Red and green dots mean a center of a circle. The misalignment error occurs due to inaccurate intrinsic parameters. Our simulator produces misaligned focal stack images because of the hardware limitations for autofocus.} 
    \label{fig:experiment}
    \vspace{-0.1cm}
\end{figure}

\noindent\textbf{Simulator.}\quad
   Because public datasets do not describe changes in FoVs or hardware limitations in off-the-shelf cameras, we propose a useful simulator to render realistic sequential defocus images for training our alignment network. Here, the most important part is to determine the error ranges of the intrinsic camera parameters, such as principal points and focal distances. We estimate them as the following process in~\Figref{fig:experiment}: (1) We capture circle patterns on a flat surface using various smartphone models by changing focus distances. (2) We initially align focal stacks with the recorded focus distances. (3) After the initial alignment, we decompose the residual motions of the captured circles using 3 basis vectors, image crop and $x-$ and $y-$axis translations. (4) We statistically calculate the error ranges of the principal points and focus distances from the three parameters of the basis vectors. Given metadata of cameras used, our simulator renders focal stacks induced from blur scales based on the focus distance and the error ranges of the basis vector.

\subsection{Focal Stack-oriented Feature Extraction}
\label{sec:feature_extraction}
For high-quality depth prediction, we consider three requirements that must be imposed on our network.
First, to robustly measure focus in the feature space, it is effective to place a gap space in the convolution operations, as proved in~\cite{jeon2019ring}. In addition, even though feature downsampling such as a convolution with strides and pooling layers is necessary to reduce the computations in low-level computer vision tasks like stereo matching~\cite{mayer2016large}, such downsampling operations can make a defocused image and its feature map sharper. This fails to accurately determine the focus within the DfF framework. Lastly, feature representations for DfF need to identify subtle distinctions in blur magnitudes between input images. 
  \begin{figure}[ht]
    \includegraphics[width=1\linewidth]{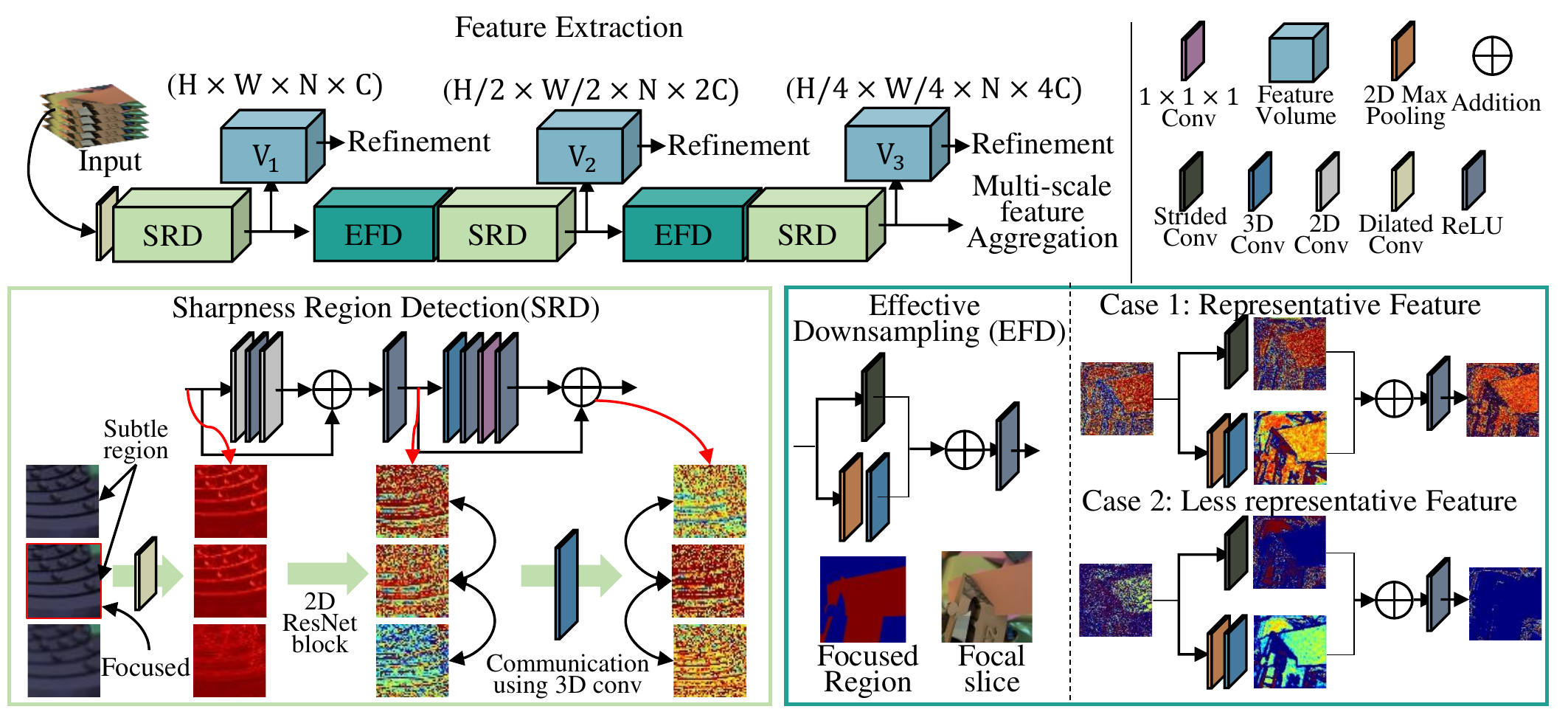}
    \vspace{-0.7cm}
    \caption{An architecture of our feature extraction. If feature maps from neighbor focal slices have similar values, our SRD gives an attention score to the sharpest focal slice. Our EFD preserves informative defocus feature representation during downsampling.} 
    \label{fig:FeatureExtraction}
    \vspace{-0.1cm}
\end{figure}

\noindent\textbf{Initial feature extraction.}\quad
 In an initial feature extraction step, we utilize a dilated convolution to extract focus features. After the dilated convolution, we extract feature pyramids to refine the focal volumes in the refinement step. 
 Given an input focal stack $S\in R ^{H*W*N*3}$ where $H$, $W$ and $N$ denote the height, width and the number of focal slices respectively, we extract three pyramidal feature volumes whose size is $H/2^{L}\times W/2^{L} \times N \times C*2^{L}$ where $L \in \{ 0,1,2\}$ and $C$ is the number of channels in the focal volume. This pyramidal feature extraction consists of three structures in which SRD and EFD are iteratively performed, as described in~\Figref{fig:FeatureExtraction}. Each pyramidal feature volume is then used as the input to the next EFD module. 
 The last one is utilized as an input of the multi-scale feature aggregation step in~\secref{sec:refinement}.
 
 \noindent\textbf{Sharp Region Detector.}\quad
 The initial feature of each focal slice is needed to communicate with other neighboring focal slices, to measure the focus of the pixel of interest. A work in~\cite{maximov2020focus} extracts focus features using a global pooling layer as a communication tool across a stack dimension. However, we observe that the global pooling layer causes a loss of important information due to its inherent limitation that all values across focal slices become single values. 
 
 Using our SRD module consisting of both a 2D convolution and a 3D convolution, we overcome the limitation. In~\Figref{fig:FeatureExtraction} (left), we extract features using a 2D ResNet block and add an attention score which is computed from them by 3D convolutions and a ReLU activation. The 3D convolution enables the detection of subtle defocus variations in weakly texture-less regions by communicating the features with neighbor focal slices.
 With this module, our network encodes more informative features for the regions than previous works~\cite{maximov2020focus,Wang-ICCV-2021}.
 \noindent\textbf{EFfective Downsampling.}\quad
 Unlike stereo matching networks that use convolutions with strides for downsampling features~\cite{chang2018pyramid,shen2021cfnet}, the stride of a convolution causes a loss in spatial information because most of the focused regions may not be selected. 
 As a solution to this issue, one previous DfF work~\cite{maximov2020focus} uses a combination of maxpooling and average pooling with the feature extraction step. 
 
 Inspired by \cite{maximov2020focus}, we propose a EFD module leveraging a well-known fact that a feature has higher activation in a focused region than weakly textured regions in~\Figref{fig:FeatureExtraction} (right). The EFD module employs a 2D max-pooling as a downsampling operation and applies a 3D convolution to its output. Through our EFD module, our network can both take representative values of focused regions in a local window and communicate the focal feature with neighbor focal slices.

\subsection{Aggregation and Refinement}
\label{sec:refinement}
Our network produces a final depth map after multi-scale feature aggregation and refinement steps.

\noindent\textbf{Multi-scale feature aggregation.}\quad
  The receptive field of our feature extraction module might be too small to learn non-local features. Therefore, we propose a multi-scale feature aggregation module using one hour-glass module to expand the receptive field, which is similar to the stereo matching network in~\cite{shen2021cfnet}. 
  At an initial step, we use three different sizes of kernels (2$\times$2, 4$\times$4, 8$\times$8) in the average pooling layer. Unlike~\cite{shen2021cfnet}, the reason for using average pooling is to avoid a memory consumption issue because DfF requires more input images.
  We then apply a ResBlock on each output of average pooling in order to extract multi-scale features. These features are embedded into the encoder and aggregated by the decoder of the hour-glass module. The aggregated feature volume is utilized as an input in the refinement step.

\noindent\textbf{Refinement and Regression.}\quad
The refinement module has three hour-glass modules with skip-connections like~\cite{chang2018pyramid}. Here, we add transposed convolutions to resize the output of each hourglass whose size is the same as each level of a pyramidal feature volume from the feature extraction module. We construct an input focal volume of each hourglass by concatenating pyramidal feature volumes of the feature extraction module with the output focal volume of the previous hourglass. As each hourglass handles increasingly higher resolutions with pyramidal feature volumes, the focal volumes are refined in a coarse to fine manner. 
To obtain a depth map from the output focal volumes, we multiply a focus distance value and the probability of each focus distance leading to maximal sharpness. The probability is computed by applying a normalized soft-plus in the output focal volumes in a manner similar to~\cite{Wang-ICCV-2021}. 
The whole depth prediction network is optimized using a weighted loss function $L_{depth}$ from scratch as follows:
 \begin{gather}
 \label{eq:depth loss}
    L_{depth} =\sum_{i=1}^4 w_{i} * ||D_{i}- D_{gt}||_2 
 \end{gather} 
  where $||\cdot||_2 $ means a $l_2$ loss and $D_{gt}$ indicates a ground truth depth map. $i\in\{1,2,3,4\}$ means the scale level of the hour-glass module. In our implementation, we set $w_{i}$ to 0.3, 0.5, 0.7 and 1.0, respectively.

\noindent\textbf{Implementation details.}\quad
 We train our network using the following strategy: (1) We first train the alignment network in~\secref{sec:alignment} during 100 epochs using the alignment loss in~\eqnref{eq:alignment loss}. (2) We freeze the alignment network and merge it with the depth prediction network. (3) We train the merged network during 1500 epochs with the depth loss in~\eqnref{eq:depth loss}. (4) In an inference step, we can estimate the depth map from the misaligned focal stack in an end-to-end manner. We note that our network is able to use an arbitrary number of images as input, like the previous CNN-based DfF/DfD~\cite{maximov2020focus,Wang-ICCV-2021}. The number of parameters of our alignment network and feature extraction module is 0.195M and 0.067M, respectively. And, the multi-scale feature aggregation module and the refinement module have 2.883M and 1.067M learnable parameters, respectively. That's, the total parameters of our network is 4.212M.
 We implement our network using a public PyTorch framework~\cite{paszke2019pytorch}, and optimize it using Adam optimizer~\cite{kingma2014adam} ($\beta_1 = 0.9, \; \beta_2 = 0.99$) with a learning rate $10^{-3}$. Our model is trained on a single NVIDIA RTX 2080Ti GPU with 4 mini-batches, which usually takes three days. For data augmentation, we apply random spatial transforms (rotation, flipping and cropping) and color jittering (brightness, contrast and gamma correction).

\section{Evaluation}
We compare the proposed network with state-of-the-art methods related to DfD, DfF and depth from light field images. We also conduct extensive ablation studies to demonstrate the effectiveness of each component of the proposed network. 
For quantitative evaluation, we use standard metrics as follows: mean absolute error (MAE), mean squared error (MSE), absolute relative error (AbsRel), square relative error (SqRel), root mean square error (RMSE), log root-mean-squared error (RMSE log), bumpiness (Bump), inference time (Secs) and accuracy metric $\delta_i = 1.25^i$ for $i \in\{1,2,3\}$. Following~\cite{Wang-ICCV-2021}, we exclude pixels whose depth ranges are out of focus distance at test time.

\subsection{Comparisons to State-of-the-art Methods}
We validate the robustness of the proposed network by showing experimental results on various public datasets: DDFF 12-Scene~\cite{hazirbas2018deep}, DefocusNet Dataset~\cite{maximov2020focus}, 4D Light Field Dataset~\cite{honauer2016dataset}, Smartphone~\cite{herrmann2020learning} as well as focal stack images generated from our simulator. The datasets provide pre-aligned defocused images. We use the training split of each dataset to build our depth estimation network in both \secref{sec:feature_extraction} and \secref{sec:refinement} from scratch, and validate it on the test split.
\begin{table}[t]
\caption{Quantitative evaluation on DDFF 12-Scene~\cite{hazirbas2018deep}. We directly refer to the results from \cite{Wang-ICCV-2021}. Since the result of DefocusNet~\cite{maximov2020focus} is not uploaded in the official benchmark, we only bring the MAE value from~\cite{maximov2020focus}. \textbf{bold}: Best, \underline{Underline}: Second best. Unit: pixel.}
\centering
\small

\resizebox{\linewidth}{!}{
\begin{tabular}{l|cccccccc}

    Method~~ & \multicolumn{1}{c}{~~MSE~$\downarrow$~~} & \multicolumn{1}{c}{~RMSE log~$\downarrow$~} & \multicolumn{1}{c}{~~AbsRel~$\downarrow$~~} & \multicolumn{1}{c}{~~SqRel~$\downarrow$~~} &\multicolumn{1}{c}{~~Bump~$\downarrow$~~}   &{~~$\delta = 1.25$~$\uparrow$~~} & {~~$\delta =1.25^{2}$~$\uparrow$~~} &{~~$\delta=1.25^{3}$~$\uparrow$~~}\\ \hline

      Lytro  & \multicolumn{1}{c}{$~~2.1e^{-3}~~$} & \multicolumn{1}{c}{~~0.31~~} & \multicolumn{1}{c}{~~0.26~~} & \multicolumn{1}{c}{\bf{~~0.01~~}} & \multicolumn{1}{c}{~~1.0~~} & \multicolumn{1}{c}{~~55.65~~} & \multicolumn{1}{c}{~~82.00~~} & \multicolumn{1}{c}{~~93.09~~}  \\ 

      VDFF~\cite{moeller2015variational} & \multicolumn{1}{c}{$~~7.3e^{-3}~~$} & \multicolumn{1}{c}{~~1.39~~} & \multicolumn{1}{c}{~~0.62~~} & \multicolumn{1}{c}{~~0.05~~} & \multicolumn{1}{c}{~~0.8~~} & \multicolumn{1}{c}{~~8.42~~} & \multicolumn{1}{c}{~~19.95~~} & \multicolumn{1}{c}{~~32.68~~}  \\  
      
      PSP-LF~\cite{zhao2017pyramid}  & \multicolumn{1}{c}{$~~2.7e^{-3}~~$} & \multicolumn{1}{c}{~~0.45~~} & \multicolumn{1}{c}{~~0.46~~} & \multicolumn{1}{c}{~~\underline{0.03}~~} & \multicolumn{1}{c}{\bf{~~0.5~~}} & \multicolumn{1}{c}{~~39.70~~} & \multicolumn{1}{c}{~~65.56~~} & \multicolumn{1}{c}{~~82.46~~}  \\ 

      PSPNet~\cite{zhao2017pyramid}  & \multicolumn{1}{c}{$~~9.4e^{-4}~~$} & \multicolumn{1}{c}{~~\underline{0.29}~~} & \multicolumn{1}{c}{~~0.27~~} & \multicolumn{1}{c}{\bf{~~0.01~~}} & \multicolumn{1}{c}{~~\underline{0.6}~~} & \multicolumn{1}{c}{~~62.66~~} & \multicolumn{1}{c}{~~85.90~~} & \multicolumn{1}{c}{~~\underline{94.42}~~}  \\ 
      
      DFLF~\cite{hazirbas2018deep}  & \multicolumn{1}{c}{$~~4.8e^{-3}~~$} & \multicolumn{1}{c}{~~0.59~~} & \multicolumn{1}{c}{~~0.72~~} & \multicolumn{1}{c}{~~0.07~~} & \multicolumn{1}{c}{~~0.7~~} & \multicolumn{1}{c}{~~28.64~~} & \multicolumn{1}{c}{~~53.55~~} & \multicolumn{1}{c}{~~71.61~~}  \\ 

      DDFF~\cite{hazirbas2018deep}   & \multicolumn{1}{c}{$~~9.7e^{-4}~~$} & \multicolumn{1}{c}{~~0.32~~} & \multicolumn{1}{c}{~~0.29~~} & \multicolumn{1}{c}{\bf{~~0.01~~}} & \multicolumn{1}{c}{~~\underline{0.6}~~} & \multicolumn{1}{c}{~~61.95~~} & \multicolumn{1}{c}{~~85.14~~} & \multicolumn{1}{c}{~~92.98~~}  \\ 

      DefocusNet~\cite{maximov2020focus}  & \multicolumn{1}{c}{~$9.1e^{-4}$~} & \multicolumn{1}{c}{~~-~~} & \multicolumn{1}{c}{~~-~~} & \multicolumn{1}{c}{~~-~~} & \multicolumn{1}{c}{~~-~~} & \multicolumn{1}{c}{~~-~~} & \multicolumn{1}{c}{~~-~~} & \multicolumn{1}{c}{~~-~~}  \\ 

      AiFDepthNet~\cite{Wang-ICCV-2021}   & \multicolumn{1}{c}{$\underline{~8.6e^{-4}~}$} & \multicolumn{1}{c}{~~\underline{0.29}~~} & \multicolumn{1}{c}{~~\underline{0.25}~~} & \multicolumn{1}{c}{}{\bf{~~0.01~~}} & \multicolumn{1}{c}{~~\underline{0.6}~~} & \multicolumn{1}{c}{~~\underline{68.33}~~} & \multicolumn{1}{c}{~~\underline{87.40}~~} & \multicolumn{1}{c}{~~93.96~~}  \\ 
      
      Ours& \multicolumn{1}{c}{$\bf{~5.7e^{-4}~}$} & \multicolumn{1}{c}{\bf{~~0.21~~}} & \multicolumn{1}{c}{\bf{~~0.17~~}} & \multicolumn{1}{c}{\bf{~~0.01~~}} & \multicolumn{1}{c}{~~\underline{0.6}~~} & \multicolumn{1}{c}{\bf{~~77.96~~}} & \multicolumn{1}{c}{\bf{~~93.72~~}} & \multicolumn{1}{c}{\bf{~~97.94~~}} \\ 

      \hline
      
\end{tabular}}
 \vspace{-0.3cm}
 \label{tab:DDFFbenchmark}
 \end{table}

\noindent\textbf{DDFF 12-Scene~\cite{hazirbas2018deep}.}\quad
DDFF 12-Scene dataset provides focal stack images and its ground truth depth maps captured by a light-field camera and a RGB-D sensor, respectively. 
The images have shallow DoFs and show texture-less regions. 
Our method shows the better performance than those of recent published works in~\tabref{tab:DDFFbenchmark} and achieves the top rank in almost evaluation metrics of the benchmark site\footnote{\url{https://competitions.codalab.org/competitions/17807\#results}}.

\begin{table}[t]
\caption{Quantitative evaluation on DefocusNet dataset~\cite{maximov2020focus}~(unit: meter), 4D Light Field dataset~\cite{honauer2016dataset}~(unit: pixel) and Smartphone dataset~\cite{herrmann2020learning}~(unit: meter). For DefocusNet dataset and 4D Light Field dataset, we directly refer to the results from \cite{Wang-ICCV-2021}. For Smartphone dataset~\cite{herrmann2020learning}, we multiply confidence scores on metrics ('MAE' and 'MSE') which are respectively denoted as 'MAE*' and 'MSE*'. \textbf{bold}: Best.}
\centering
\small

\resizebox{\linewidth}{!}{
\begin{tabular}{l|ccc|ccc|ccc}
 ~~ & \multicolumn{3}{c|}{~~DefocusNet Dataset~\cite{maximov2020focus}~~}  & \multicolumn{3}{c|}{~~4D Light Field~\cite{honauer2016dataset}~~}& \multicolumn{3}{c}{~~Smartphone~\cite{herrmann2020learning}~~} \\ 
  Method~~ & \multicolumn{1}{c}{~~MAE~$\downarrow$~~} & \multicolumn{1}{c}{~~MSE~$\downarrow$~~} & \multicolumn{1}{c|}{~~AbsRel~$\downarrow$~~} & \multicolumn{1}{c}{~~MSE~$\downarrow$~~} & \multicolumn{1}{c}{~~RMSE~$\downarrow$~~} &\multicolumn{1}{c|}{~~Bump~$\downarrow$~~}  & \multicolumn{1}{c}{~~MAE*~$\downarrow$~~} &\multicolumn{1}{c}{~~MSE*~$\downarrow$~~}&\multicolumn{1}{c}{~~Secs~$\downarrow$~~}\\ \hline
  DefocusNet~\cite{maximov2020focus}  & \multicolumn{1}{c}{~~0.0637~~} & \multicolumn{1}{c}{~~0.0175~~} & \multicolumn{1}{c|}{~~0.1386~~} & \multicolumn{1}{c}{~~0.0593~~} & \multicolumn{1}{c}{~~0.2355~~} &   \multicolumn{1}{c|}{~~2.69~~} & \multicolumn{1}{c}{~~0.1650~~} & \multicolumn{1}{c}{~~0.0800~~} & \multicolumn{1}{c}{~~0.1598~~} \\ 
  AiFDepthNet~\cite{Wang-ICCV-2021}   & \multicolumn{1}{c}{~~0.0549~~} & \multicolumn{1}{c}{~~0.0127~~} & \multicolumn{1}{c|}{~~0.1115~~} & \multicolumn{1}{c}{~~0.0472~~} & \multicolumn{1}{c}{~~0.2014~~} &
  \multicolumn{1}{c|}{~~1.58~~}& \multicolumn{1}{c}{~~0.1568~~} & \multicolumn{1}{c}{~~0.0764~~}
  & \multicolumn{1}{c}{~~0.1387~~}\\
  Ours& \multicolumn{1}{c}{\bf{~~0.0403~~}} & \multicolumn{1}{c}{\bf{~~0.0087~~}} & \multicolumn{1}{c|}{\bf{~~0.0809~~}} &  \multicolumn{1}{c}{\bf{~~0.0230~~}} & \multicolumn{1}{c}{\bf{~~0.1288~~}} &   \multicolumn{1}{c|}{\bf{~~1.29~~}}& \multicolumn{1}{c}{\bf{~~0.1394~~}} & \multicolumn{1}{c}{\bf{~~0.0723~~}}& \multicolumn{1}{c}{\bf{~~0.1269~~}} \\ \hline
 \end{tabular}}
 
 \vspace{-0.2cm}

 \label{tab:table1}
 \end{table}

  \begin{figure}[tb!]
    \includegraphics[width=1\linewidth]{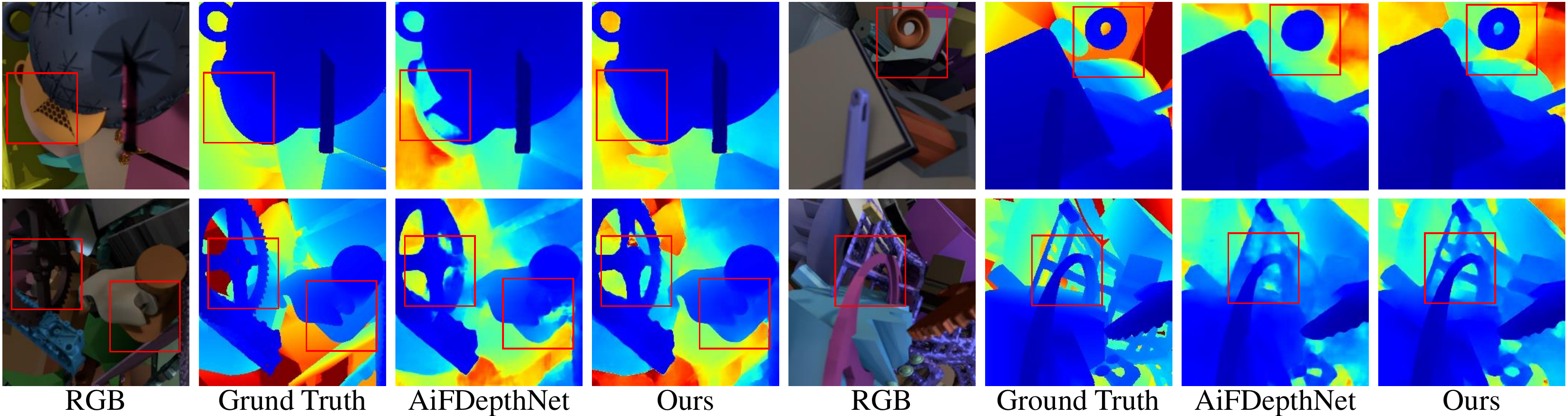}
    \vspace{-0.7cm}
    \caption{Examples of depth prediction from AiFDepthNet and ours on DefocusNet dataset.} 
    \label{fig:Defocus}
\end{figure}

\begin{figure}[t!]
    \includegraphics[width=1\linewidth]{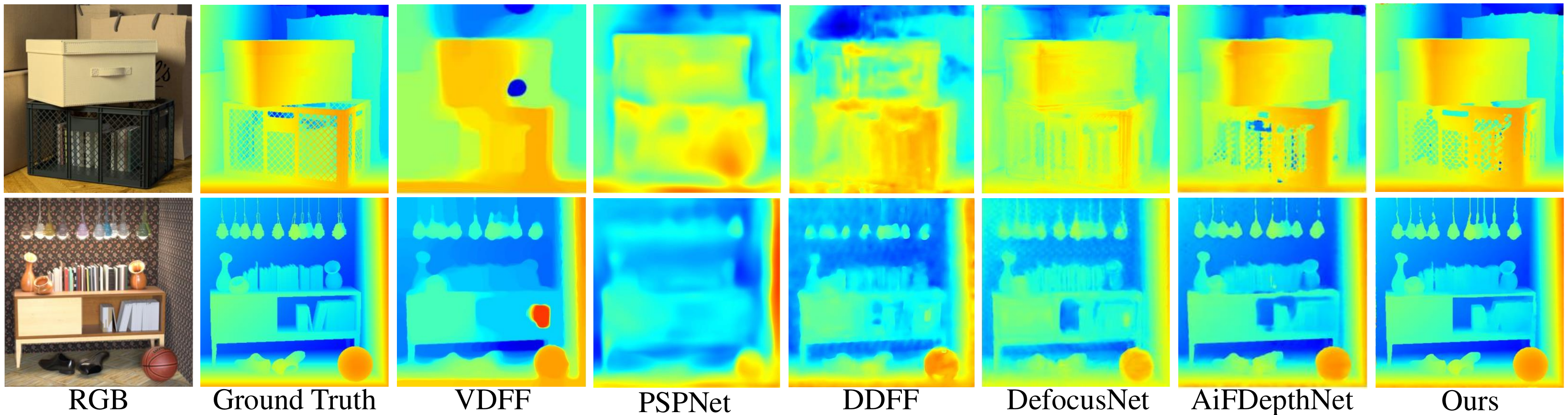}
    \vspace{-0.7cm}
    \caption{Qualitative comparison on 4D Light Field dataset.
    }
    \vspace{-0.7cm}

\label{fig:4D_LF}
\end{figure}

\noindent\textbf{DefocusNet Dataset~\cite{maximov2020focus}.}\quad
This dataset is rendered in a virtual space and generated using Blender Cycles renderer~\cite{blender2018blender}. 
Focal stack images consist of only five defocused images whose focus distances are randomly sampled in an inverse depth space. The quantitative results are shown in \tabref{tab:table1}. As shown in~\Figref{fig:Defocus}, our method successfully reconstructs the smooth surface and the sharp depth discontinuity rather than previous methods.

\noindent\textbf{4D Light Field Dataset~\cite{honauer2016dataset}.}\quad
This synthetic dataset has 10 focal slices with shallow DoFs for each focal stack. The number of focal stacks in training and test split is 20 and 4, respectively. For fair comparison on this dataset, we follow the evaluation protocol in the relevant work~\cite{Wang-ICCV-2021}.
In qualitative comparisons~\Figref{fig:4D_LF}, our SRD and EFD enable to capture sharp object boundaries like the box and fine details like lines hanging from the ceiling. In quantitative evaluation of~\tabref{tab:table1} the MSE and RMSE are half of them from the comparison methods~\cite{abuolaim2020defocus,wanner2012globally}.

\begin{figure}[t!]
    \includegraphics[width=1\linewidth]{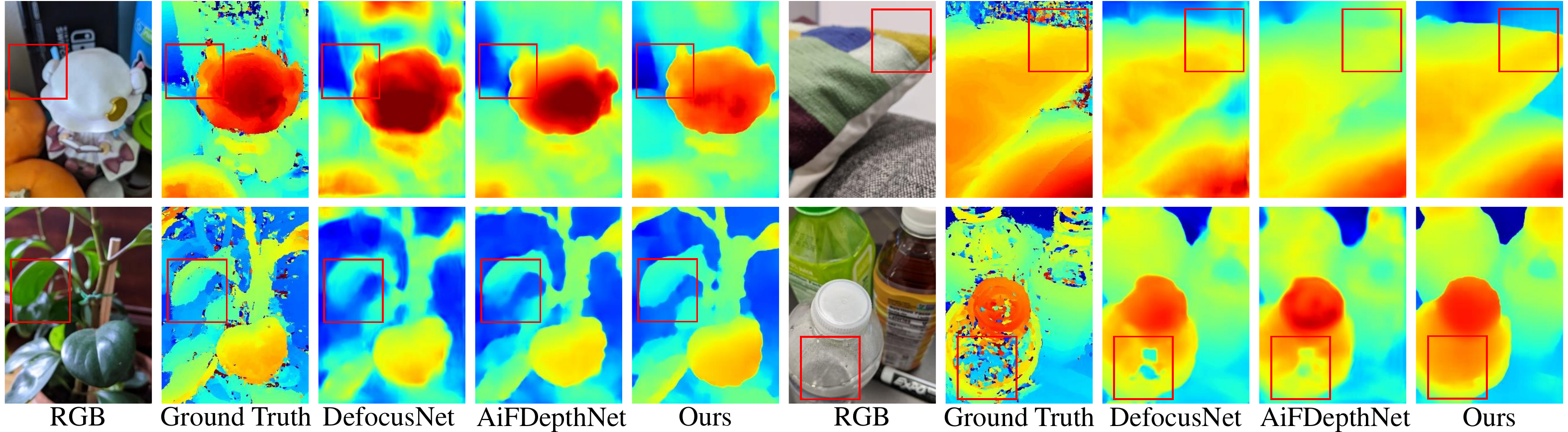}
    \vspace{-0.7cm}
    \caption{ Qualitative results on Smartphone dataset.}
    \vspace{-0.1cm}

    \label{fig:Smartphone}
\end{figure}

\noindent\textbf{Smartphone~\cite{herrmann2020learning}.}\quad
This dataset shows real-world scenes captured from Pixel 3 smartphones. Unlike previous datasets, ground truth depth maps are obtained by multiview stereo~\cite{schoenberger2016sfm,schoenberger2016mvs} and its depth holes are not considered in the evaluation. 
As expected, our network achieves the promising performance over the state-of-the-art methods, whose results are reported in \tabref{tab:table1} and \Figref{fig:Smartphone}. 
We note that our method consistently yields the best quality depth maps from focal stack images regardless of dataset, thanks to our powerful defocused feature representations using both SRD and EFD.

 \begin{table}[t]
 \caption{Quantitative result across different datasets for generalization of the state-of-the-art methods and ours. We train our depth prediction model on FlyingThings3D and test them on Middlebury stereo~(unit: pixel) and DefocusNet dataset~(unit: meter). For fair comparison, we directly refer the results of the works~\cite{Wang-ICCV-2021,maximov2020focus} from \cite{Wang-ICCV-2021}.}
\centering
\small
\resizebox{\linewidth}{!}{
\begin{tabular}{l|cc|cccccc}
\multicolumn{1}{l|}{Method~~}  & \multicolumn{1}{c}{~~Train Dataset~~} & \multicolumn{1}{c|}{~~Test Dataset~~} & \multicolumn{1}{c}{~~MAE~$\downarrow$~~} & \multicolumn{1}{c|}{~~MSE~$\downarrow$~~} & \multicolumn{1}{c}{~~RMSE~$\downarrow$~~} & \multicolumn{1}{c}{~~AbsRel~$\downarrow$~~} & \multicolumn{1}{c}{~~SqRel~$\downarrow$~~}\\ \hline
   DefocusNet~\cite{maximov2020focus} & \multicolumn{1}{c}{~~~~} & \multicolumn{1}{c|}{~~~~} & \multicolumn{1}{c}{~~7.408~~} & \multicolumn{1}{c}{~~157.440~~} & \multicolumn{1}{c}{~~9.079~~} & \multicolumn{1}{c}{~~0.231~~} & \multicolumn{1}{c}{~~4.245~~}  \\ 
  AiFDepthNet~\cite{Wang-ICCV-2021}  & \multicolumn{1}{c}{~~FlyingThings3d~~} & \multicolumn{1}{c|}{~~Middlebury~~} & \multicolumn{1}{c}{~~3.825~~} & \multicolumn{1}{c}{~~58.570~~} & \multicolumn{1}{c}{~~5.936~~} & \multicolumn{1}{c}{~~0.165~~} & \multicolumn{1}{c}{~~3.039~~} \\

  Ours& \multicolumn{1}{c}{~~~~}& \multicolumn{1}{c|}{~~~~} & \multicolumn{1}{c}{\bf{~~1.645~~}} & \multicolumn{1}{c}{\bf{~~9.178~~}} & \multicolumn{1}{c}{\bf{~~2.930~~}} & \multicolumn{1}{c}{\bf{~~0.068~~}} & \multicolumn{1}{c}{\bf{~~0.376~~}} \\ \hline
  DefocusNet~\cite{maximov2020focus}  & \multicolumn{1}{c}{~~~~}& \multicolumn{1}{c|}{~~~~} & \multicolumn{1}{c}{~~0.320~~} & \multicolumn{1}{c}{~~0.148~~} & \multicolumn{1}{c}{~~0.372~~} & \multicolumn{1}{c}{~~1.383~~} & \multicolumn{1}{c}{~~0.700~~}  \\ 
  AiFDepthNet~\cite{Wang-ICCV-2021}   & \multicolumn{1}{c}{~~FlyingThings3d~~}& \multicolumn{1}{c|}{~~DefocusNet~~} & \multicolumn{1}{c}{~~0.183~~} & \multicolumn{1}{c}{~~0.080~~} & \multicolumn{1}{c}{~~0.261~~} & \multicolumn{1}{c}{~~0.725~~} & \multicolumn{1}{c}{~~0.404~~}  \\
  Ours& \multicolumn{1}{c}{~~~~}& \multicolumn{1}{c|}{~~~~} & \multicolumn{1}{c}{\bf{~~0.163~~}} &
\multicolumn{1}{c}{\bf{~~0.076~~}} & \multicolumn{1}{c}{\bf{~~0.259~~}} & \multicolumn{1}{c}{\bf{~~0.590~~}} & \multicolumn{1}{c}{\bf{~~0.360~~}} \\ \hline

\end{tabular}}

 \vspace{-0.4cm}
 \label{tab:cross_domain}
 \end{table}
 \begin{figure}[t!]
    \includegraphics[width=1\linewidth]{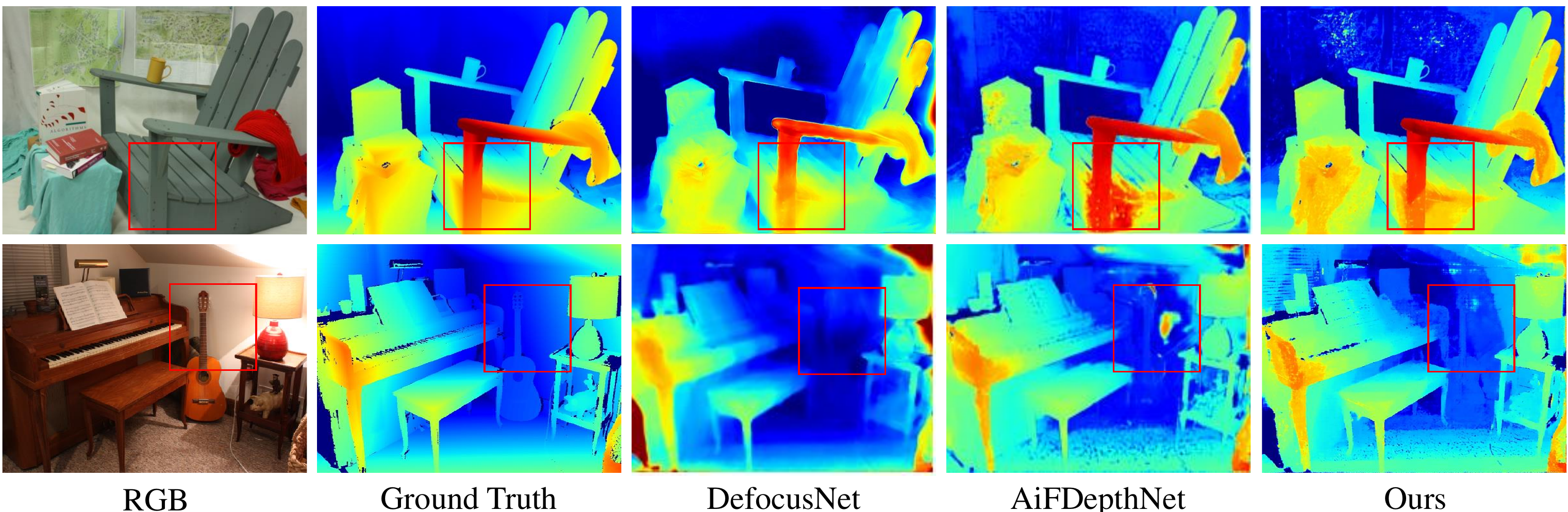}
    \vspace{-0.7cm}
    \caption{ Qualitative results on Middlebury dataset.  
} \label{fig:Middlebury}
\vspace{-0.7cm}
\end{figure}

\noindent\textbf{Generalization across different datasets.}\quad
Like~\cite{Wang-ICCV-2021}, we demonstrate the generality of the proposed network. For this, we train our network on Flyingthings3D~\cite{mayer2016large} which is a large-scale synthetic dataset, and test it on two datasets including Middlebury Stereo~\cite{scharstein2014high} and DefocusNet dataset~\cite{maximov2020focus}. As shown in \tabref{tab:cross_domain} and \Figref{fig:Middlebury}, our network still shows impressive results on both datasets.

\begin{figure}[t]
    \includegraphics[width=1\linewidth]{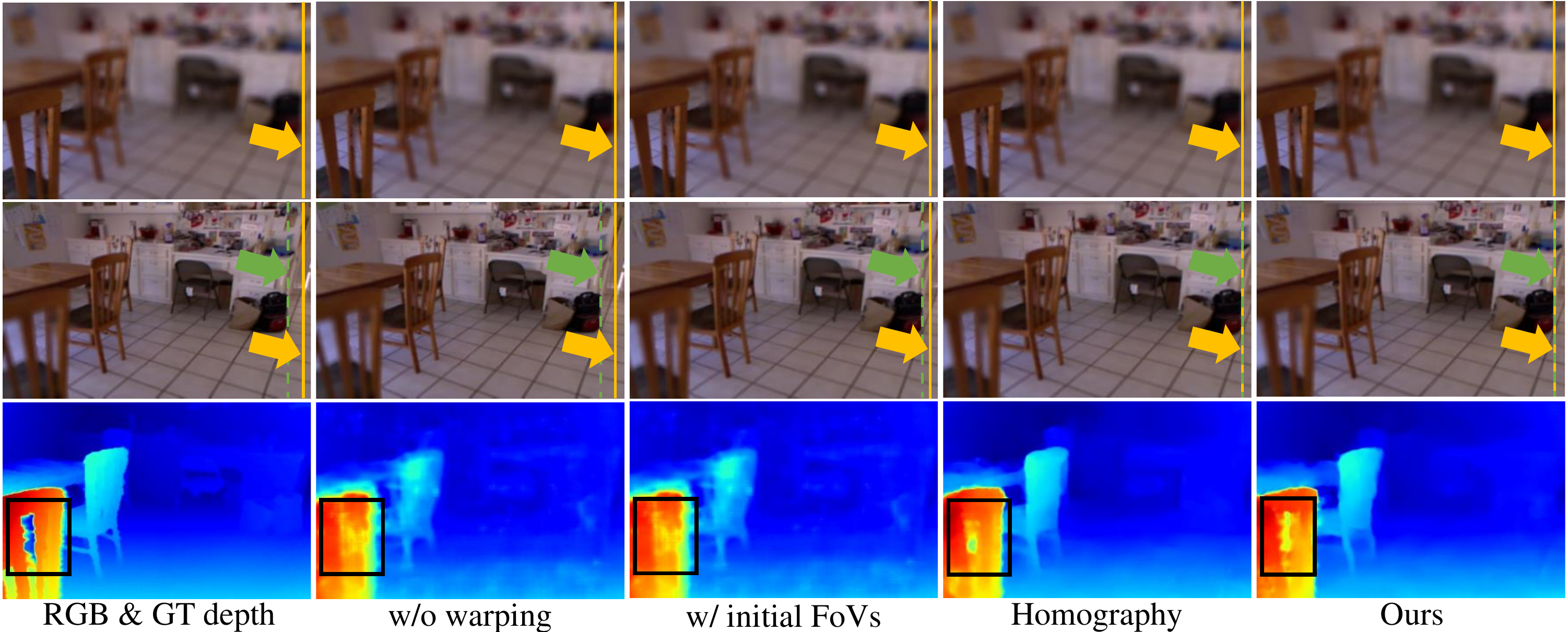}
    \vspace{-0.7cm}
    \caption{Ablation study on our alignment network. The first and second row refer a target and reference focal slice whose FoVs have the smallest and the biggest values, respectively. The third row shows depth estimation results in accordance to the alignment methods.} 
    \label{fig:ablation_alignment}
    \vspace{-0.2cm}

\end{figure}

\begin{table}[t]
\caption{Ablation study for alignment network. Unit: meter
 } 
\centering
\small

\resizebox{\linewidth}{!}{
\begin{tabular}{l|cccccccccc}

    Module~~ & \multicolumn{1}{c}{~~MAE~$\downarrow$~~} & \multicolumn{1}{c}{~~MSE~$\downarrow$~~} &\multicolumn{1}{c}{~~RMSE log~$\downarrow$~~} & \multicolumn{1}{c}{~~AbsRel~$\downarrow$~~} & \multicolumn{1}{c}{~~SqRel~$\downarrow$~~}    &{~~$\delta = 1.25$~$\uparrow$~~} & {~~$\delta =1.25^{2}$~$\uparrow$~~} &{~~$\delta=1.25^{3}$~$\uparrow$~~}&{~~Secs~$\downarrow$~~}&{~~GPU~~~~}\\ \hline
      w/o alignment& \multicolumn{1}{c}{~~0.0247~~} & \multicolumn{1}{c}{~~0.0014~~} & \multicolumn{1}{c}{~~0.0915~~} & \multicolumn{1}{c}{~~0.0067~~} & \multicolumn{1}{c}{~~0.0034~~} & \multicolumn{1}{c}{~~0.9707~~} & \multicolumn{1}{c}{~~0.9970~~} & \multicolumn{1}{c}{~~0.9995~~} &
      \multicolumn{1}{c}{~~\bf{0.0107}~~}&\multicolumn{1}{c}{2080Ti}\\ 

      w/ initial FoVs & 
      \multicolumn{1}{c}{~~0.0165~~} & \multicolumn{1}{c}{~~0.0009~~} &  \multicolumn{1}{c}{~~0.0636~~} &\multicolumn{1}{c}{~~0.0400~~} & \multicolumn{1}{c}{~~0.0019~~}  & \multicolumn{1}{c}{~~0.9867~~} & \multicolumn{1}{c}{~~0.9976~~} & \multicolumn{1}{c}{~~0.9994~~} & \multicolumn{1}{c}{~~0.0358~~} &\multicolumn{1}{c}{2080Ti}\\

      Homography-based~~~~~~~~~~~~~~~~~~& \multicolumn{1}{c}{~~\bf{0.0151}~~} & \multicolumn{1}{c}{~~\bf{0.0007}~~} & \multicolumn{1}{c}{~~\bf{0.0570}~~} & \multicolumn{1}{c}{~~0.0369~~} & \multicolumn{1}{c}{~~\bf{0.0015}~~} & \multicolumn{1}{c}{~~\bf{0.9907}~~} & \multicolumn{1}{c}{~~\bf{0.9986}~~} & \multicolumn{1}{c}{~~\bf{0.9997}~~} & \multicolumn{1}{c}{~~0.8708~~}&\multicolumn{1}{c}{R3600} \\ \hline
      
      Ours& \multicolumn{1}{c}{~~\bf{0.0151}~~} & \multicolumn{1}{c}{~~\bf{0.0007}~~} & \multicolumn{1}{c}{~~0.0578~~} & \multicolumn{1}{c}{~~\bf{0.0365}~~} & \multicolumn{1}{c}{~~0.0016~~} & \multicolumn{1}{c}{~~0.9898~~} & \multicolumn{1}{c}{~~0.9984~~} & \multicolumn{1}{c}{~~0.9996~~} &\multicolumn{1}{c}{~~0.0923~~}&\multicolumn{1}{c}{2080Ti}\\  \hline 
    \end{tabular}}
 
 \vspace{-0.5cm}
 \label{tab:AlignmentNetwork}
 \end{table}

 \begin{figure}[t]
    \includegraphics[width=1\linewidth]{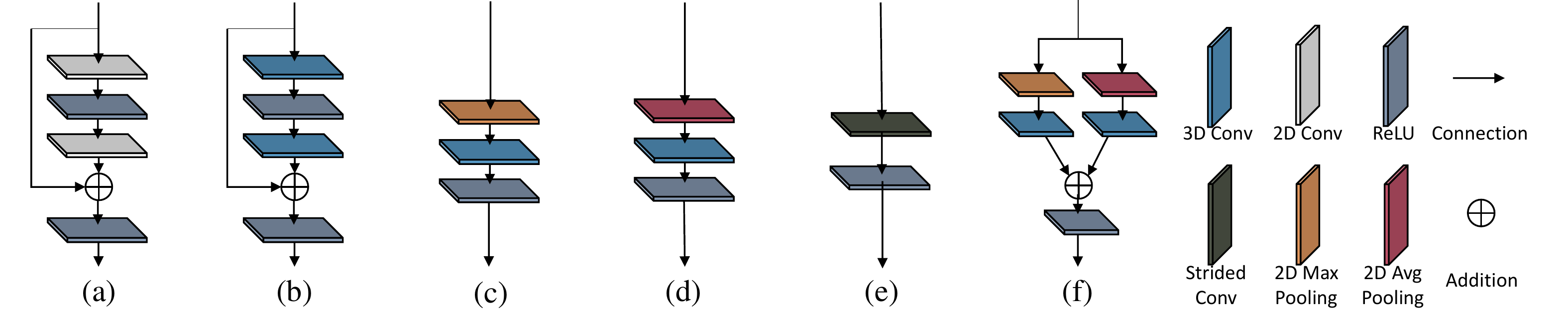}
    \vspace{-0.9cm}
    \caption{Candidate modules of our SRD and EFD. (a) 2D ResNet block, (b) 3D ResNet block, (c) Max pooling + 3D Conv, (d) Average pooling + 3D Conv, (e) Strided Conv and (f) 3D pooling layer.
}
\label{fig:ablation_SRD_EFD}
\end{figure}

\begin{table}[ht]
 \caption{Ablation studies for SRD and EFD. Unit: meter
 }
\centering
\small
\resizebox{\linewidth}{!}{
\begin{tabular}{l|cccccccc}

    Module~~ & \multicolumn{1}{c}{~~MAE~$\downarrow$~~} & \multicolumn{1}{c}{~~MSE~$\downarrow$~~} &\multicolumn{1}{c}{~~RMSE log~$\downarrow$~~} & \multicolumn{1}{c}{~~AbsRel~$\downarrow$~~} & \multicolumn{1}{c}{~~SqRel~$\downarrow$~~}    &{~~$\delta = 1.25$~$\uparrow$~~} & {~~$\delta =1.25^{2}$~$\uparrow$~~} &{~~$\delta=1.25^{3}$~$\uparrow$~~}\\ \hline
      SRD ~$\rightarrow$~ 2D ResNet block& \multicolumn{1}{c}{~~0.0421~~} & \multicolumn{1}{c}{~~0.0095~~} & \multicolumn{1}{c}{~~0.1614~~} & \multicolumn{1}{c}{~~0.0842~~} & \multicolumn{1}{c}{~~0.0142~~} & \multicolumn{1}{c}{~~0.9082~~} & \multicolumn{1}{c}{~~0.9722~~} & \multicolumn{1}{c}{~~0.9873~~}  \\ 

      SRD ~$\rightarrow$~ 3D ResNet block& \multicolumn{1}{c}{~~0.0409~~} & \multicolumn{1}{c}{~~0.0088~~} &  \multicolumn{1}{c}{~~0.1576~~} &\multicolumn{1}{c}{~~0.0818~~} & \multicolumn{1}{c}{~~\bf{0.0128}~~}  & \multicolumn{1}{c}{~~0.9123~~} & \multicolumn{1}{c}{~~0.9725~~} & \multicolumn{1}{c}{~~0.9891~~}  \\ \hline

      EFD ~$\rightarrow$~ Maxpooling~+~3D Conv& \multicolumn{1}{c}{~~0.0421~~} & \multicolumn{1}{c}{~~0.0094~~} & \multicolumn{1}{c}{~~0.1622~~} & \multicolumn{1}{c}{~~0.0845~~} & \multicolumn{1}{c}{~~0.0143~~} & \multicolumn{1}{c}{~~0.9125~~} & \multicolumn{1}{c}{~~0.9712~~} & \multicolumn{1}{c}{~~0.9849~~}  \\ 
      
      EFD ~$\rightarrow$~ Avgpooling~+~3D Conv& \multicolumn{1}{c}{~~0.0422~~} & \multicolumn{1}{c}{~~0.0097~~} & \multicolumn{1}{c}{~~0.1628~~} & \multicolumn{1}{c}{~~0.0830~~} & \multicolumn{1}{c}{~~0.0141~~} & \multicolumn{1}{c}{~~0.9126~~} & \multicolumn{1}{c}{~~0.9718~~} & \multicolumn{1}{c}{~~0.9860~~}  \\ 
      
      EFD ~$\rightarrow$~ Strided Conv& \multicolumn{1}{c}{~~0.0419~~} & \multicolumn{1}{c}{~~0.0091~~} & \multicolumn{1}{c}{~~0.1630~~} & \multicolumn{1}{c}{~~0.0842~~} & \multicolumn{1}{c}{~~0.0135~~} & \multicolumn{1}{c}{~~0.9144~~} & \multicolumn{1}{c}{~~0.9725~~} & \multicolumn{1}{c}{~~0.9867~~}  \\ 
      
      EFD ~$\rightarrow$~ 3D Poolying Layer& \multicolumn{1}{c}{~~0.0414~~} & \multicolumn{1}{c}{~~0.0089~~} & \multicolumn{1}{c}{~~0.1594~~} & \multicolumn{1}{c}{~~0.0843~~} & \multicolumn{1}{c}{~~0.0132~~} & \multicolumn{1}{c}{~~0.9088~~} & \multicolumn{1}{c}{~~0.9747~~} & \multicolumn{1}{c}{~~0.9886~~}  \\ \hline
        Ours   & \multicolumn{1}{c}{\bf{~~0.0403~~}} & \multicolumn{1}{c}{\bf{~~0.0087~~}} & \multicolumn{1}{c}{\bf{~~0.1534~~}} & \multicolumn{1}{c}{\bf{~~0.0809~~}} & \multicolumn{1}{c}{{~~0.0130~~}} & \multicolumn{1}{c}{\bf{~~0.9137~~}} & \multicolumn{1}{c}{\bf{~~0.9761~~}} & \multicolumn{1}{c}{\bf{~~0.9900~~}}  \\ \hline
\end{tabular}}
 \vspace{-0.3cm}
 \label{tab:submoduleAblation}
 \end{table}

  \begin{figure}[ht!]
    \includegraphics[width=1\linewidth]{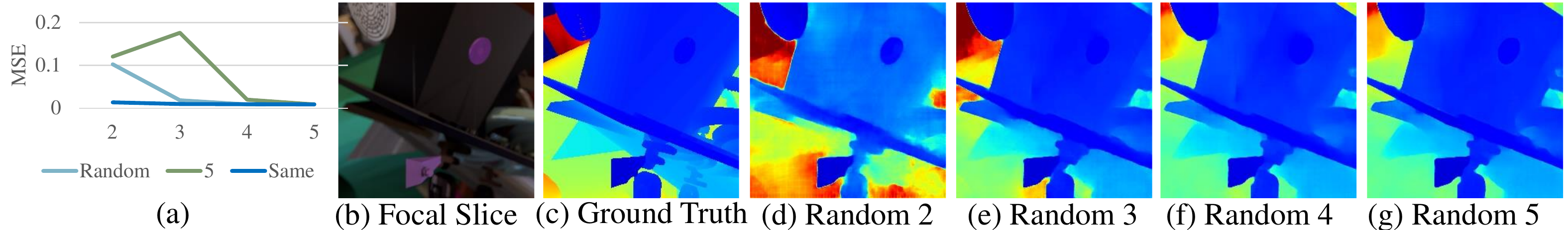}
    \vspace{-0.8cm}
    \caption{ (a) The performance change according to the number of focal slices in training and test phase. (b) One of focal slices and (c) its ground truth depth map. (d) to (g) output depth maps on the random number of input focal slices in training phase. 
} \label{fig:Random_input}
 \vspace{-0.5cm}

\end{figure}

\subsection{Ablation studies}
We carry out extensive ablation studies to demonstrate the effectiveness of each module of the proposed network.

\noindent\textbf{Alignment network.}\quad
  We first evaluate our alignment network. To do this, we render focal stacks using our simulator which generates defocused images based on a camera metadata. We test our alignment network in consideration of four cases: 1) without any warping, 2) with only initial FoVs in \eqref{eq:Fov}, 3) a classical homography method~\cite{evangelidis2008parametric}, 4) our alignment network with initial FoVs. The quantitative results are reported in \tabref{tab:AlignmentNetwork}, whose example is displayed in~\Figref{fig:ablation_alignment}. The results demonstrate that our alignment network achieves much faster and competable performance with the classic homography-based method.

\noindent\textbf{SRD and EFD.}\quad
We compare our modules with other feature extraction modules depicted in \Figref{fig:ablation_SRD_EFD}. 
We conduct this ablation study on DefocusNet dataset~\cite{maximov2020focus} because it has more diverse DoF values than other datasets.
The quantitative result is reported in \tabref{tab:submoduleAblation}.  

When we replace our SRD module with either 3D ResNet block or 2D ResNet block only, there are performance drops, even with more learnable parameters for the 3D ResNet block.
We also compare our EFD module with four replaceable modules: max-pooling+3D Conv, average pooling+3D Conv, Stride convolution and 3D pooling layer. 
As expected, our EFD module achieves the best performance because it allows better gradient flows preserving defocus property. 

\noindent\textbf{Number of focal slices.}\quad
 Like previous DfF networks \cite{Wang-ICCV-2021,maximov2020focus}, our network can handle an arbitrary number of focal slices by the virtue of 3D convolutions. Following the relevant work~\cite{Wang-ICCV-2021}, we train our network from three different ways, whose result is reported in~\Figref{fig:Random_input}: The '5' means a model trained using five focal slices; The 'Same'
 denotes that the number of focal slices in training and test phase is same; The 'Random' is a model trained using an arbitrary number of focal slices.
 
  The '5' case performs poorly when the different number of focal slices is used in the test phase, and the 'Same' case shows promising performances. Nevertheless, the 'Random' case consistently achieves good performances regardless of the number of focal slices

\vspace{-0.3cm}
\section{Conclusion}
\vspace{-0.2cm}
 In this paper, we have presented a novel and true end-to-end DfF architecture. To do this, we first propose a trainable alignment network for sequential defocused images. We then introduce a novel feature extraction and an efficient downsampling module for robust DfF tasks. The proposed network achieves the best performance in the public DfF/DfD benchmark and various evaluations.
 
 \noindent\textbf{Limitation.}\quad There are still rooms for improvements. A more sophisticated model for flow fields in the alignment network would enhance depth prediction results. More parameters can be useful for extreme rotations.
 Another direction is to make depth prediction better by employing focal slice selection like defocus channel attention in the aggregation process.
 
\noindent\textbf{Acknowledgement}\quad
 This work is in part supported by the Institute of Information $\&$ communications Technology Planning $\&$ Evaluation (IITP) (No.2021-0-02068, Artificial Intelligence Innovation Hub), Vehicles AI Convergence Research $\&$ Development Program through the National IT Industry Promotion Agency of Korea (NIPA), `Project for Science and Technology Opens the Future of the Region' program through the INNOPOLIS FOUNDATION (Project Number: 2022-DD-UP-0312) funded by the Ministry of Science and ICT (No.S1602-20-1001), the National Research Foundation of Korea (NRF) ( No. 2020R1C1C10\\12635) grant funded by the Korea government (MSIT), the Ministry of Trade, Industry and Energy (MOTIE) and Korea Institute for Advancement of Technology (KIAT) through the International Cooperative R$\&$D program (P0019797), and the GIST-MIT Collaboration grant funded by the GIST in 2022.
\clearpage
%
%
\bibliographystyle{splncs04}
\bibliography{egbib}
\end{document}